\documentclass[a4paper,11pt]{article}

\usepackage{graphicx}  
\usepackage{url}       
\usepackage{times}
\usepackage{natbib}
\usepackage{amsmath}
\usepackage{amssymb}
\usepackage[margin=25mm]{geometry}
\usepackage{xcolor}
\usepackage[normalem]{ulem}

\usepackage{tcolorbox}
\usepackage{hyperref}
\usepackage{graphicx}
\usepackage{geometry}
\usepackage{setspace} 
\title{InSpatio-WorldFM: An Open-Source Real-Time Generative Frame Model}

\date{}

\author{
\centering
\begin{tabular}{c}
\textbf{InSpatio Team (alphabetical order)}\\[0.5em]
\textbf{D}onghui Shen, \textbf{G}uofeng Zhang, \textbf{H}aomin Liu, \textbf{H}aoyu Ji, \textbf{J}ialin Liu, \textbf{J}ing Guo,\\
\textbf{N}an Wang, \textbf{S}iji Pan, \textbf{W}eihong Pan, \textbf{W}eijian Xie, \textbf{X}iaojun Xiang, \textbf{X}iaoyu Zhang,\\
\textbf{X}ianbin Liu, \textbf{Y}ifu Wang, \textbf{Y}ipeng Chen, \textbf{Z}hewen Le, \textbf{Z}hichao Ye, \textbf{Z}iqiang Zhao
\end{tabular}
}

\begin{document}
\maketitle

\vspace{-8mm}
\begin{tcolorbox}[
    colback=white,
    colframe=white,
    arc=10pt,
    boxrule=1.5pt,
    left=0pt,
    right=0pt,
    top=0pt,
    bottom=0pt,
]

We present InSpatio-WorldFM, an open-source real-time frame model for spatial intelligence. Unlike video-based world models that rely on sequential frame generation and incur substantial latency due to window-level processing, InSpatio-WorldFM adopts a frame-based paradigm that generates each frame independently, enabling low-latency real-time spatial inference. By enforcing multi-view spatial consistency through explicit 3D anchors and implicit spatial memory, the model preserves global scene geometry while maintaining fine-grained visual details across viewpoint changes. We further introduce a progressive three-stage training pipeline that transforms a pretrained image diffusion model into a controllable frame model and finally into a real-time generator through few-step distillation. Experimental results show that InSpatio-WorldFM achieves strong multi-view consistency while supporting interactive exploration on consumer-grade GPUs, providing an efficient alternative to traditional video-based world models for real-time world simulation.

\vspace{10pt}

\begin{minipage}[b]{0.7\textwidth}
\textbf{Project Website:} \href{https://inspatio.github.io/worldfm/}{\textcolor{blue}{https://inspatio.github.io/worldfm/}}

\textbf{Github:} \href{https://github.com/inspatio/worldfm}{\textcolor{blue}{https://github.com/inspatio/worldfm}}

\end{minipage}%
\end{tcolorbox}

\vspace{-20pt}

\begin{figure}[!ht]
  \centering
   \includegraphics[width=\linewidth]{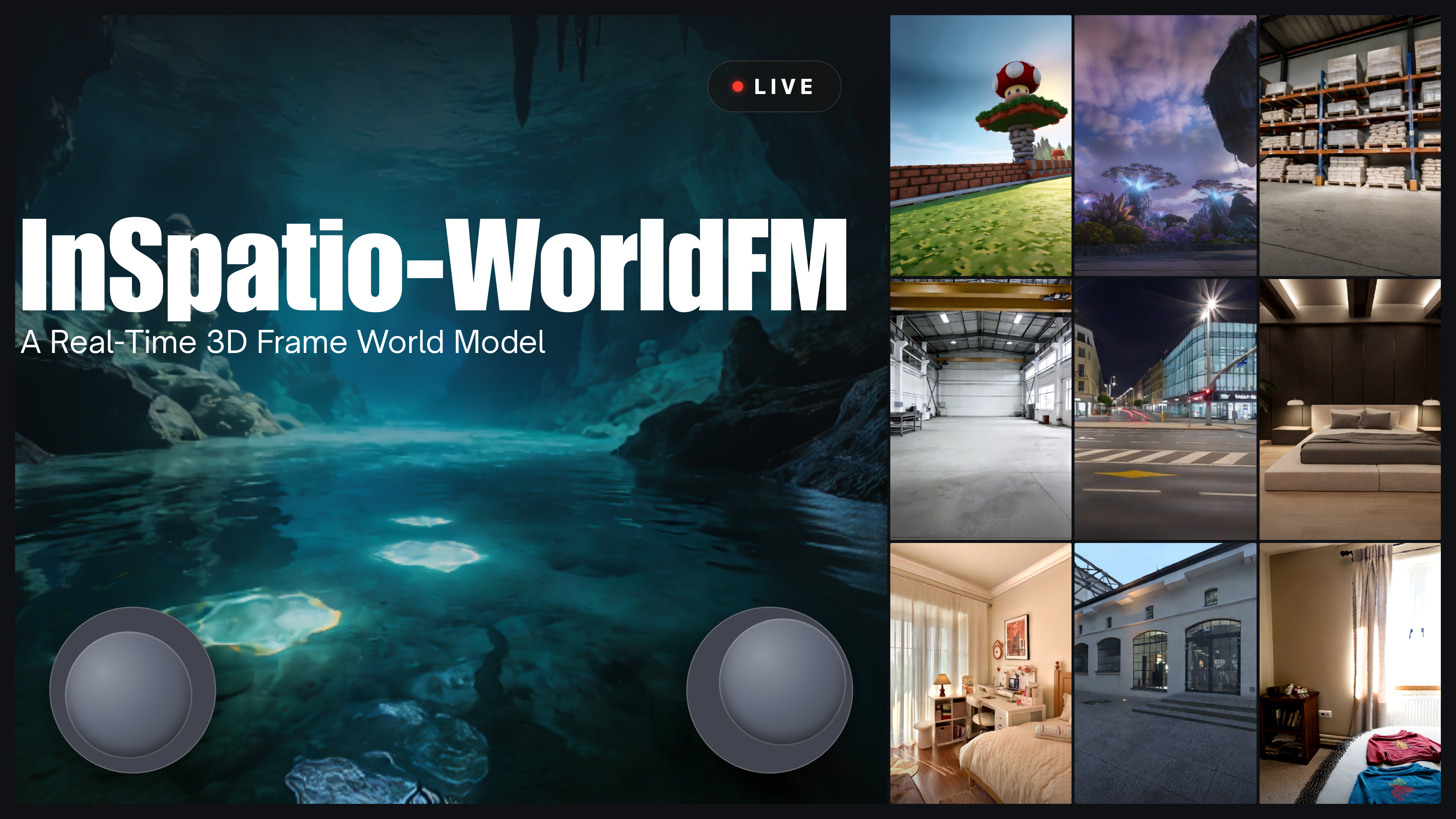}   
    \caption{Examples of generated worlds across diverse styles, including photorealistic, science-fiction, game-like, and artistic environments. The joystick interface enables real-time interactive exploration with negligible latency.}   
   \label{fig:teaser}     
\end{figure}

\newpage 

\section{Introduction}
Recent advances in generative models~\citep{rombach2022high,flux2024,labs2025flux1kontextflowmatching,kong2024hunyuanvideo,blattmann2023stable,peebles2023scalable} have significantly improved visual realism and temporal coherence. These models learn rich motion patterns and camera dynamics from large-scale video datasets~\cite{ling2024dl3dv,zhou2018stereo,yao2020blendedmvs}, which has sparked growing interest in using generative models to simulate persistent realistic environments. As a result, the research focus is gradually shifting from traditional text-to-video generation~\citep{bar2024lumiere,singer2022make,brooks2024video,yang2024cogvideox,videoworldsimulators2024,wan2025} toward the broader goal of building world models capable of representing and interacting with structured environments~\citep{zhou2025scenex,yang2025layerpano3d,chung2023luciddreamer,yu2024wonderjourney,yu2025wonderworld,gao2024cat3d}, such as HY-World~\citep{team2025hunyuanworld} and LingBot-World~\citep{lingbot-world}, as well as the models that incorporate predefined rules or constraints~\citep{liu2024physics3d,zhang2024physdreamer,raistrick2024infinigen}, such as Genie 3~\citep{genie3}.

Despite this progress, most existing world models~\citep{huang2025voyager,wu2025video,ren2025gen3c,ding2025understanding,zhang2025matrix,sun2025worldplay} are still built upon video generation architectures. Leveraging video models allows these systems to inherit strong priors about motion, appearance, and camera movement learned from large-scale video data. However, this paradigm also introduces several fundamental limitations.

\textbf{Interactive latency remains inevitable.} Most video-based world models generate frames sequentially within a temporal window. With bidirectional attention and full-window decoding, each generation step must process all frames in the window, resulting in substantial inference overhead. Even with distillation-based acceleration, this window-level dependency fundamentally limits the ability of such models to support truly real-time interaction.

\textbf{Spatial errors accumulate over time.} Video models are primarily optimized for short-term temporal continuity rather than long-term spatial consistency. As generation progresses, small spatial inaccuracies may accumulate, leading to structural drift or inconsistencies in scene geometry. Without explicit mechanisms to enforce global spatial constraints, correcting these accumulated errors remains challenging.

Recently, World Labs proposed a real-time frame model (RTFM)~\citep{rtfm2025}, exploring the potential of frame-based world modeling. However, the technical details provided are quite limited and the source code is not publicly available. 
To further explore this direction and address the limitations of video-based world models, we propose InSpatio-WorldFM, a frame-based world model designed for real-time spatial reasoning and generation.
Unlike video-based approaches that model the world as a sequence of dependent frames, InSpatio-WorldFM directly incorporates spatial structure into the generation of individual frames. By conditioning each frame on explicit spatial information, the model can maintain consistent scene geometry while enabling low-latency frame synthesis. This design allows the system to generate spatially coherent observations while supporting real-time interaction. We introduce several key components:
\begin{itemize}
\item \textbf{Multi-view consistent training data curation.} We construct training data with explicit multi-view consistency, allowing the model to learn stable spatial relationships across viewpoints.

\item \textbf{Progressive three-stage training pipeline} that evolves a foundation image generator (Stage I), into a controllable frame model with spatial memory (Stage II), and finally into a real-time few-step generator (Stage III). 

\item \textbf{Real-time generation} enabled by few-step distillation~\citep{yin2024one} through 2-step denoising. Experimental results demonstrate that InSpatio-WorldFM can generate frames with strong multi-view consistency while achieving real-time inference without interactive latency.
\end{itemize}

Together, these components enable InSpatio-WorldFM to serve as an efficient foundation for real-time spatial intelligence, providing a practical alternative to traditional video-based world models.

\begin{figure}[t]
  \centering
   \includegraphics[width=\linewidth]{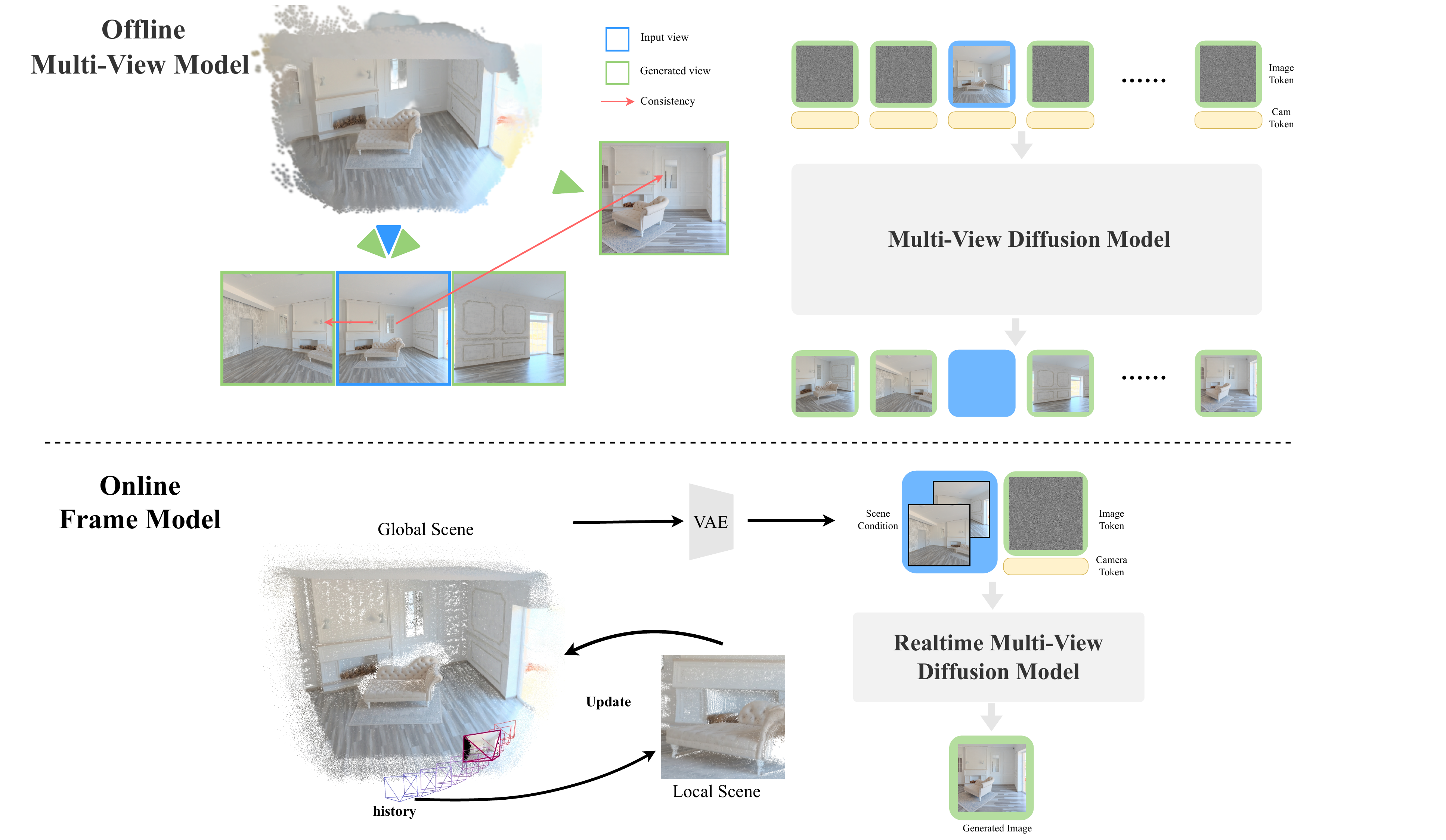}   
   \caption{\textbf{Overview.} In the offline stage, a multi-view-consistent model generates plausible observations that provide 3D anchors and reference appearances. In the online stage, frame model performs fast real-time inference while updating scene content at keyframes.}
   \label{fig:arch}   
\end{figure}

\section{InSpatio-WorldFM}
\subsection{Overview}
As illustrated in Fig.~\ref{fig:arch}, the framework of InSpatio-WorldFM consists of an offline stage and an online stage. 
In the offline stage, a single image is used as input to a multi-view consistent model (e.g.,~\citep{zhou2025stable,gao2024cat3d}) to generate multi-view consistent observations, which provide reference appearances, while the 3D anchors can be obtained using a reconstruction model (e.g., ~\citep{wang2025vggt,wang2024dust3r,wang2024moge}).
Alternatively, a simpler approach is to generate a panoramic image, which can also provide 360° scene constraints, and several open-source methods are already available for this task~\citep{team2025hunyuanworld,feng2023diffusion360,zhang2024taming}.
In the online stage, we introduce an efficient frame model that performs real-time inference and generation.

\subsection{Formulation}

We formulate InSpatio-WorldFM as a conditional generative frame model that synthesizes novel-view images of a 3D scene from a single reference image under user-defined camera motion.
Let $x_{\text{ref}} \in \mathbb{R}^{H \times W \times 3}$ denotes the reference image with its associated camera pose $\pi_{\text{ref}} = (K_{\text{ref}}, E_{\text{ref}})$, where $K$ and $E$ represent the camera intrinsic and extrinsic matrices, respectively.
Given a target camera pose $\pi_{\text{tgt}}$, our goal is to generate the corresponding target-view image $x_{\text{tgt}}$ that is geometrically consistent with $x_{\text{ref}}$.

We approach this task within a latent diffusion framework.
Let $\mathcal{E}$ and $\mathcal{D}$ denote the encoder and decoder of a pretrained variational autoencoder (VAE), respectively.
The generation is formulated as learning a conditional denoising model $\epsilon_\theta$ that reverses a forward diffusion process in the latent space $z = \mathcal{E}(x)$. 
Specifically, the model learns to predict the noise $\epsilon$ added to the target latent $z_{\text{tgt}}$ at diffusion timestep $t$:

$$\mathcal{L} = \mathbb{E}_{z_{\text{tgt}}, \epsilon \sim \mathcal{N}(0, I), t} \left[ \left\| \epsilon - \epsilon_\theta(z_t, t, \mathcal{C}) \right\|^2 \right],$$
where $z_t = \alpha_t z_{\text{tgt}} + \sigma_t \epsilon$ is the noised latent at timestep $t$ following a predefined noise schedule $(\alpha_t, \sigma_t)$, and $\mathcal{C} = \{x_{\text{ref}}, \pi_{\text{ref}}, \pi_{\text{tgt}}, \hat{x}_{\text{tgt}}\}$ represents the full condition set.
Here $\hat{x}_{\text{tgt}}$ denotes the point cloud rendering at the target viewpoint with 3D foundation models~\citep{wang2025vggt,wang2024dust3r,wang2024moge,li2018megadepth}, which serves as an explicit 3D spatial anchor.

To achieve real-time / interactive rendering, we adopt a three-stage training pipeline that progressively evolves a foundation image generator into an efficient real-time frame model:

\begin{itemize}
\item \textbf{Stage I: Pre-Training.} We establish a high-fidelity image generation prior by selecting an efficient and expressive Diffusion Transformer as our backbone. The choice of the base model is driven by both generation quality and computational efficiency, as the latter directly impacts real-time deployment feasibility. 

\item \textbf{Stage II: Middle-Training.} We transform the pretrained image generator into a controllable frame model with spatial memory. This stage involves constructing dedicated training data from real-world video datasets and synthetic environments, introducing architectural modifications for camera-conditioned generation, and developing a hybrid spatial memory mechanism that combines explicit 3D anchors with implicit neural memory.

\item \textbf{Stage III: Post-Training.} We distill the multi-step diffusion model into an efficient few-step generator through distribution matching distillation, enabling real-time / interactive rendering on consumer-grade GPUs.
\end{itemize}

\subsection{Pre-Training}

The choice of the base image generation model is critical, as it must simultaneously satisfy two requirements: high-fidelity generation with physically plausible textures and scene geometry, and computational efficiency sufficient for downstream real-time deployment.
We select PixArt-$\Sigma$~\citep{chen2024pixart} as our foundation model.
PixArt-$\Sigma$ is an efficient Diffusion Transformer (DiT)~\citep{peebles2023scalable} for text-to-image generation that achieves quality competitive with state-of-the-art models at significantly lower computational cost, providing an effective balance between generation quality and inference throughput that is well-suited as the backbone for a real-time frame model.

\subsection{Middle-Training}

The middle-training stage transforms the pretrained image diffusion model into a controllable frame model capable of generating spatially consistent and interactive visual content.
Although foundation diffusion model demonstrates strong potential in high-fidelity image synthesis, it is inherently limited to generating single, isolated images without any notion of spatial coherence or interactive controllability.
We therefore construct dedicated training data and introduce architectural modifications to endow the model with action-conditioned control and spatial consistency.

\subsubsection{Fundamental Frame Model}
To achieve true real-time performance, we adopt a minimal frame-based architecture as the online inference model, generating each frame independently. To maintain multi-view consistency across successive inferences without relying on multi-frame processing, we integrate explicit 3D anchors (point cloud rendering) with implicit spatial memory (reference frame attention), enabling the model to preserve geometric coherence while operating on individual frames.

\begin{figure}[t]
  \centering
   \includegraphics[width=\linewidth]{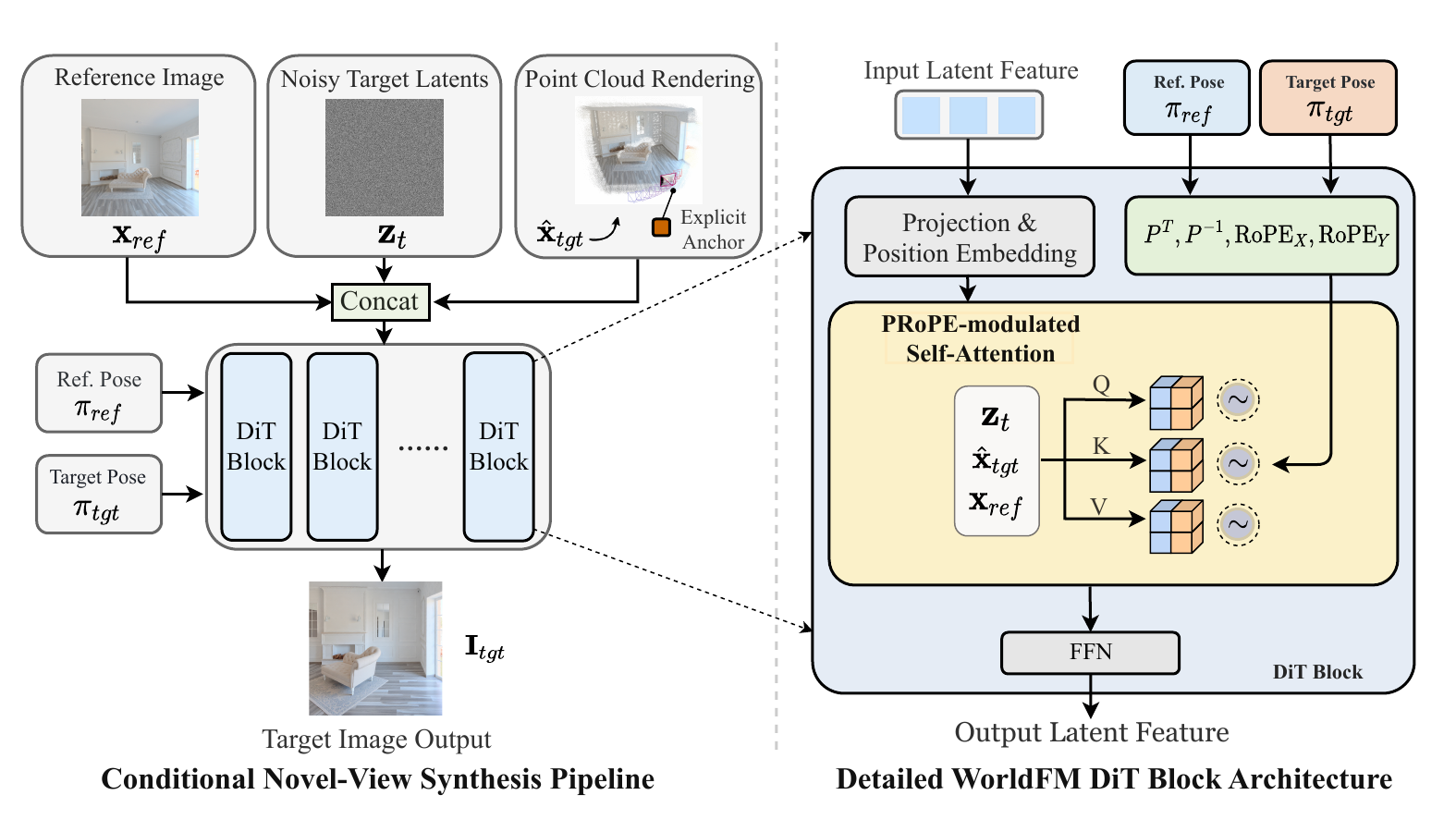}   
   \caption{\textbf{The pipeline of InSpatio-WorldFM.} The left part illustrates the conditional novel-view synthesis pipeline of WorldFM. WorldFM takes a reference image $x_{\text{ref}}$ (implicit scene memory), noisy latents $z_t$, and point cloud rendering $\hat{x}_{\text{tgt}}$ (explicit 3D anchor) as inputs, which are spatially concatenated along the width dimension. Reference pose $\pi{\text{ref}}$ and target pose $\pi_{\text{tgt}}$ are also injected as control signals. The frame-based Diffusion Transformer Blocks process these conditions and synthesizes the target view $I_{\text{tgt}}$ in real-time via Distribution Matching Distillation (DMD). The right part shows the detailed architecture of the DiT blocks in WorldFM. Camera geometry control is achieved through the Projection Relative Position Embedding (PRoPE) strategy, enhancing cross-view geometric reasoning. The hybrid spatial memory mechanism combines point cloud rendering (explicit 3D anchor) and reference image (implicit memory), interacting solely through self-attention to achieve robust 3D consistency.}   
   \label{fig:pipeline}   
\end{figure}

\noindent \textbf{Model Architecture.} InSpatio-WorldFM generates target-view images by receiving a reference image, noisy target latents, and user-defined camera motion as inputs.

We adopt a self-attention-only transformer architecture for condition injection.
Empirically, we find that injecting condition information through self-attention yields higher generation quality than cross-attention alternatives.
The input to the transformer is formed by spatially concatenating three components along the width dimension: the noisy target latent $z_t$, the point cloud rendering of the target viewpoint (cond1), and the reference frame image (cond2).
All three components are independently projected into patch tokens via a shared patch embedding layer and augmented with sinusoidal positional embeddings.
After passing through the full transformer, the output is split along the width dimension and only the target portion is retained as the final prediction.

\textbf{Camera Pose Encoding.}
Encoding camera geometry into the transformer is essential for controllable novel-view generation. We explore three strategies for injecting camera pose information:

\begin{itemize}
\item \textit{Plücker ray embedding}~\citep{he2025cameractrl,agarwal2025cosmos,alhaija2025cosmos}. For each patch token, the 6-dimensional Plücker coordinates $(\mathbf{o} \times \mathbf{d}, \mathbf{d})$ are computed in world coordinates, where $\mathbf{o}$ is the camera origin and $\mathbf{d}$ is the ray direction through the patch center. These features are projected to the hidden dimension via a two-layer MLP and added to the patch embeddings. This approach provides an explicit per-token geometric prior but encodes camera information additively without directly modulating the attention computation.

\item \textit{Projected Relative Positional Encoding (PRoPE)}~\citep{li2025cameras}. PRoPE integrates camera geometry directly into the attention mechanism by applying camera-dependent linear transformations to the query, key, and value tensors. Specifically, given camera projection matrices $P_i$ for each view $i$, PRoPE applies $P_i^\top$ to queries and $P_i^{-1}$ to keys/values, combined with 2D rotary position embeddings for intra-image spatial structure. This formulation enables the attention mechanism to reason about cross-view geometric correspondences natively.

\item \textit{Pure parametric injection}~\citep{bai2025recammaster}. Camera pose parameters (rotation and translation matrices) are directly mapped to token-level embeddings through a learned MLP and added to the hidden representations, without imposing explicit geometric structure such as ray directions or projection matrices.
\end{itemize}

We adopt PRoPE for InSpatio-WorldFM, as it exhibits the fastest convergence and the most stable camera control among the three strategies in our experiments. Compared with Plücker ray embedding, which encodes camera information additively, PRoPE directly modulates the attention computation through camera-dependent linear transformations, enabling the attention mechanism to reason about cross-view geometric correspondences natively.

\noindent\textbf{Hybrid Spatial Memory.}
A core challenge in maintaining spatial consistency across generated frames is preserving a coherent memory of the 3D scene.
RTFM~\citep{rtfm2025} relies on posed frames as primitive spatial memory, while StarGen~\citep{zhai2025stargen} warps features extracted from posed keyframes to provide spatial conditioning.
To achieve more persistent and stable spatial memory, we adopt a hybrid design that integrates explicit 3D anchors with implicit neural memory:

\begin{itemize}
\item \textit{Explicit anchors.} For each target viewpoint, a point cloud rendering and the corresponding target camera pose provide stable 3D geometric constraints, serving as global 3D priors that anchor the generation in 3D space. While point clouds effectively maintain coarse geometric structures, they are complemented by implicit memory to preserve fine-grained appearance details.

\item \textit{Implicit memory.} The reference frame and its associated camera pose supply appearance information from a previously observed viewpoint. The transformer attends to these tokens through the self-attention mechanism, allowing the model to implicitly retrieve and transfer relevant visual content to the target view.
\end{itemize}

By combining both forms of spatial memory, InSpatio-WorldFM achieves robust consistency: explicit anchors maintain coarse geometric structures and provide global 3D priors, while implicit memory preserves fine-grained appearance details and enables the model to hallucinate plausible content in unobserved regions.

\noindent \textbf{Training Data.}
We construct training pairs from three sources: publicly available videos (e.g., internet videos and video datasets such as DL3DV~\citep{ling2024dl3dv}, RealEstate10K~\citep{zhou2018stereo}), our own captured videos, and the synthetic data generated using Unreal Engine (UE)~\citep{ue}.
For each real video clip, we randomly sample 16 frames and apply a feedforward reconstruction model (e.g., MapAnything~\citep{keetha2025mapanything}) to estimate per-frame camera poses and depth maps.
From these 16 frames, 4 are selected as the reference frame group to construct the global point cloud, while the remaining 12 serve as training targets.
For each target frame, the reference frame is selected as the temporally closest one from the 4-frame reference group.
The point cloud rendering at the target viewpoint is obtained by projecting the global point cloud onto the target camera plane. To ensure robustness, we apply random shuffling and masking strategies, simulating the disorder and discreteness encountered in real-world scenarios.

\noindent \textbf{Training Strategy.}
We introduce several training strategies to encourage stable geometric learning and robust spatial memory:

\begin{itemize}
\item \textbf{Noise schedule biasing.} To prioritize learning stable geometric structure over fine-grained texture details, we increase the sampling probability of high-noise timesteps during training. This biased schedule encourages the model to first capture the coarse spatial layout before refining appearance.

\item \textbf{Progressive condition injection.} We observe that simultaneously exposing the model to both the explicit anchor (point cloud rendering) and the implicit memory (reference frame) from the beginning of training leads to rapid overfitting to the explicit anchor, as it provides a much stronger and more direct signal. This causes the model to neglect the implicit memory pathway. To ensure robustness and avoid overfitting, we adopt a progressive injection strategy: in the early phase of training, only the reference frame (implicit memory) is provided as a condition, forcing the model to develop a robust ability to extract and utilize implicit spatial cues. The explicit anchor is then gradually introduced to enhance control precision.

\item \textbf{Random anchor masking.} In the later stages of training, we randomly mask the explicit anchor with a certain probability. This regularization encourages the model to maintain its ability to leverage implicit memory, preventing it from becoming overly dependent on the explicit 3D prior.
\end{itemize}

\subsubsection{Finetuning with Synthetic Data}

Although the fundamental frame model trained on real data acquires reasonable and geometrically consistent generation capabilities, the depth and pose estimates produced by feedforward reconstruction models inevitably contain errors.
These inaccuracies introduce inter-view inconsistencies in the point cloud renderings, which undermine the stability of viewpoint transitions and content persistence.

To address this, we construct a synthetic dataset using Unreal Engine~\citep{ue} that provides precise ground-truth camera poses and depth information.
Specifically, our approach begins by selecting an initial camera pose from semantically valid regions within the scene. The trajectory is then synthesized using either stochastic motion sampling or pre-defined motion templates. Throughout generation, we enforce spatial constraints through collision avoidance to guarantee viewpoint validity.
Training pairs are then constructed analogously to the real data pipeline: 4 frames serve as the reference group and 12 as targets.

We finetune the fundamental frame model on this synthetic dataset for a limited number of steps.
The controlled exposure to synthetic data is intentional --- excessive finetuning on synthetic distributions would compromise the model's ability to generate realistic appearances on natural images.
Empirically, we find that even a small amount of synthetic finetuning yields a significant improvement in the stability of camera-controlled viewpoint transitions, suggesting that the model effectively leverages the geometric precision of synthetic data to refine its spatial reasoning without losing its natural appearance priors.

\begin{figure}[!ht]
  \centering
   \includegraphics[width=\linewidth]{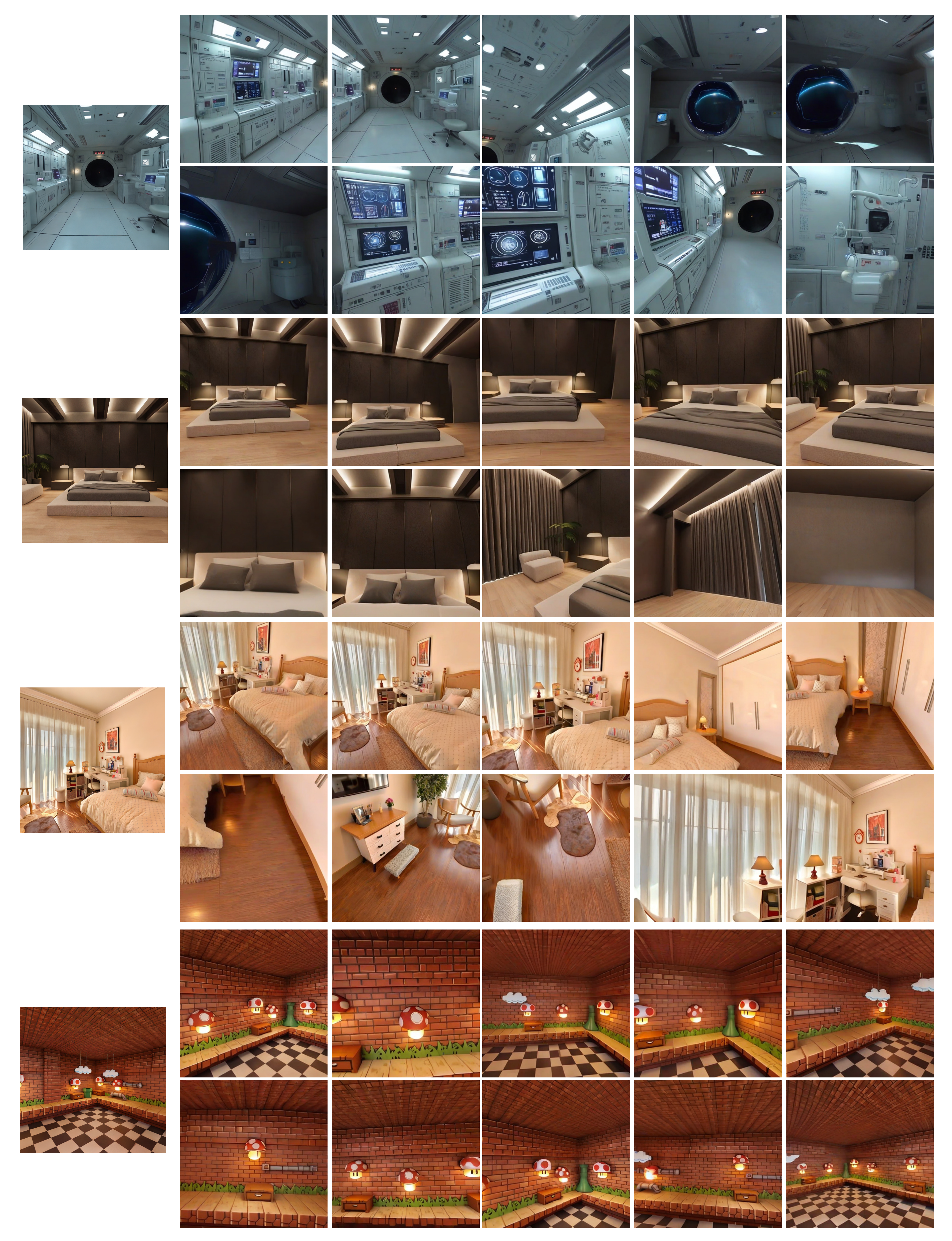}   
   \caption{Qualitative results of teacher model.}   
   \label{fig:results0} 
   \vspace{18mm}
\end{figure}
\begin{figure}[!ht]
  \centering
   \includegraphics[width=\linewidth]{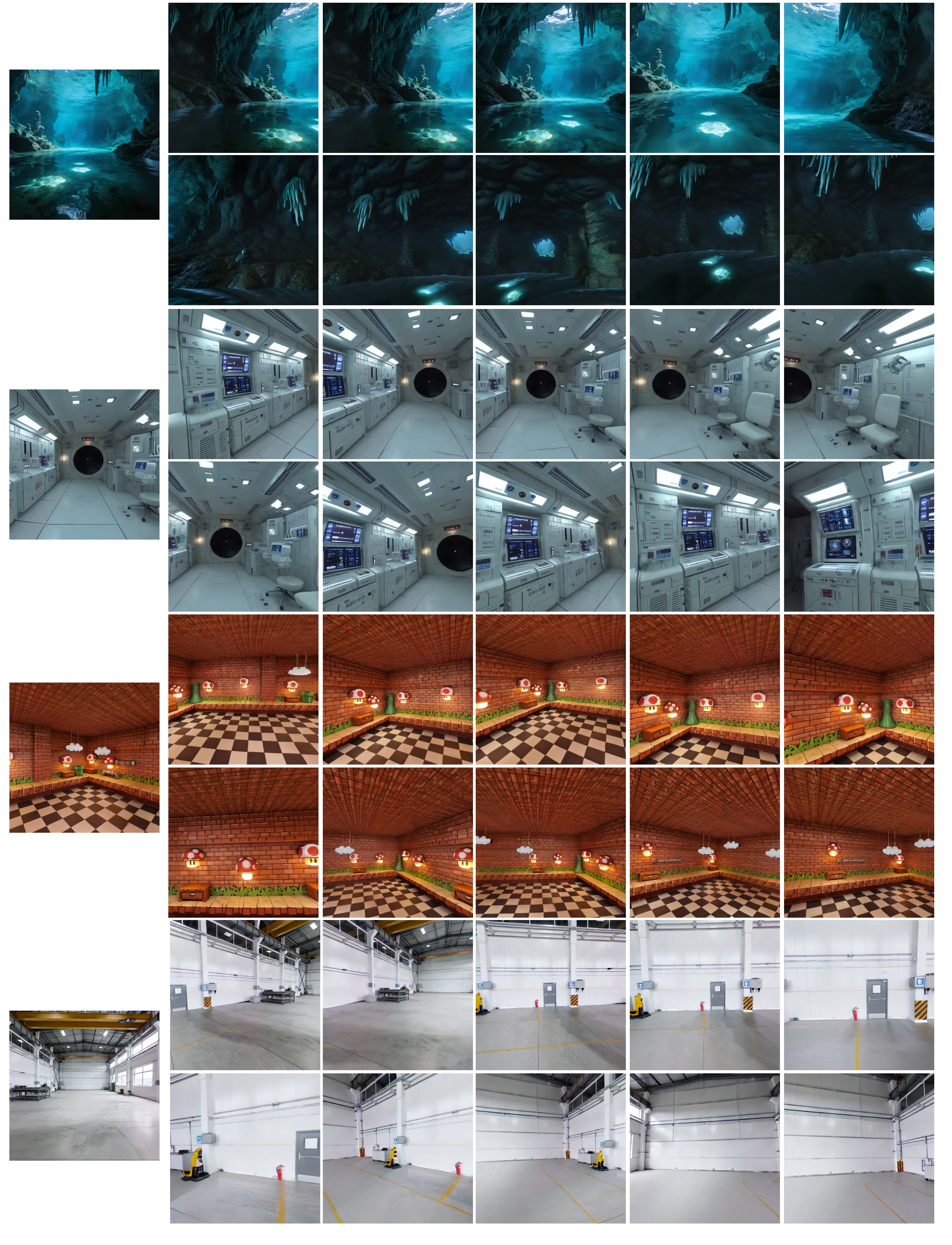}   
   \caption{Qualitative results of InSpatio-WorldFM.}   
   \label{fig:results1}   
   \vspace{18mm}
\end{figure}
\begin{figure}[!ht]
  \centering
   \includegraphics[width=\linewidth]{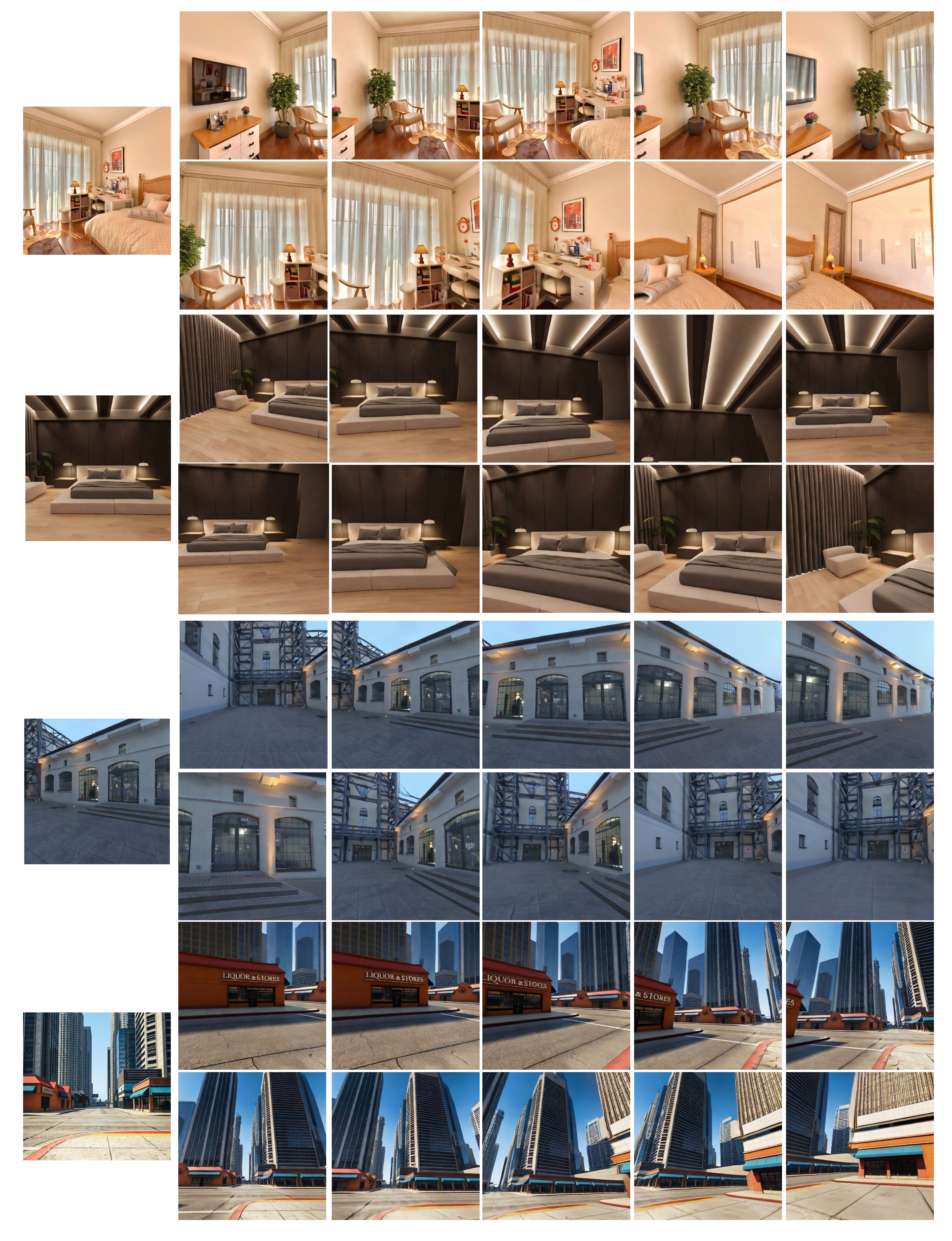}   
   \caption{Qualitative results of InSpatio-WorldFM.}   
   \label{fig:results2}   
   \vspace{18mm}
\end{figure}
\begin{figure}[!ht]
  \centering
   \includegraphics[width=\linewidth]{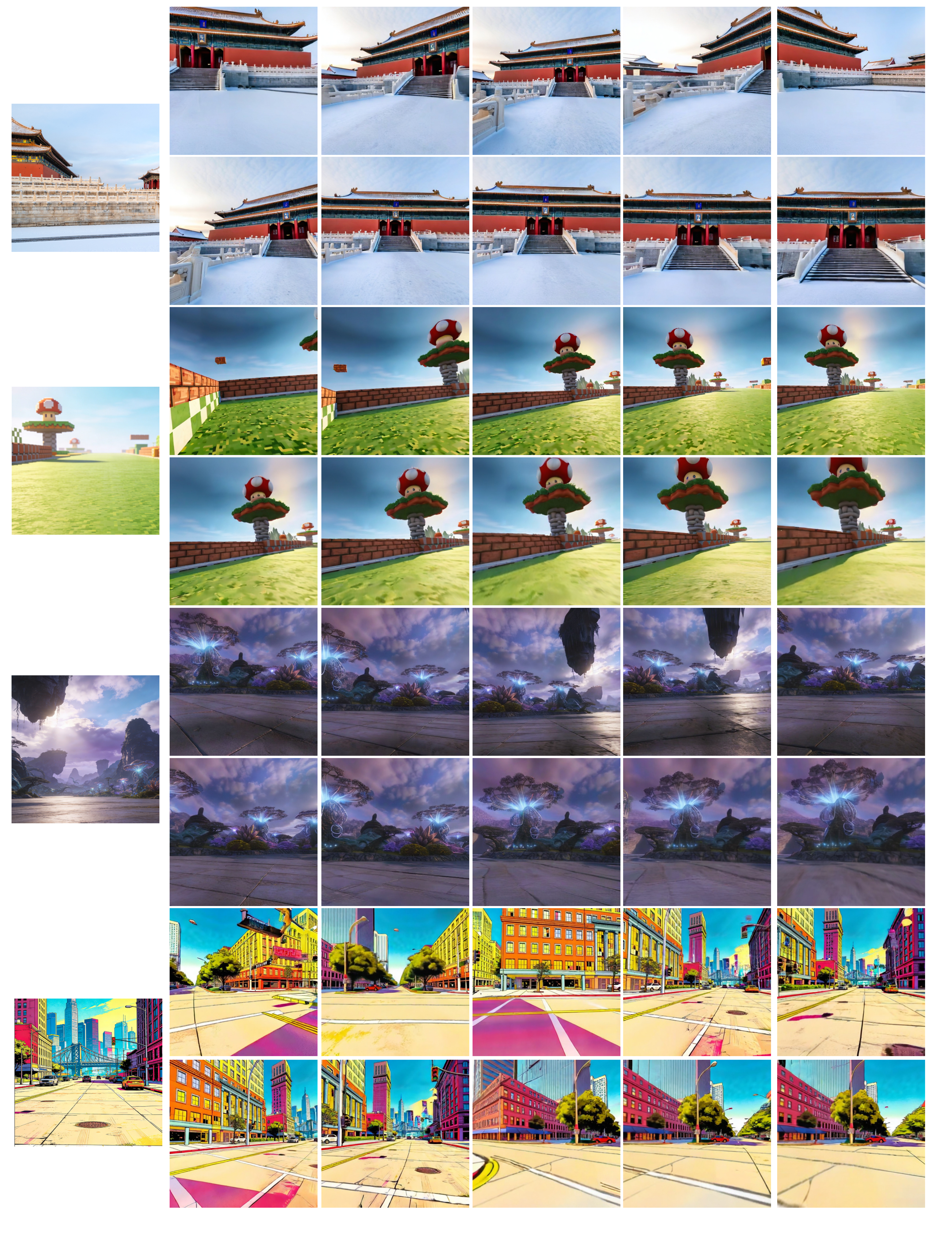}   
   \caption{Qualitative results of InSpatio-WorldFM.}   
   \label{fig:results3}   
   \vspace{18mm}
\end{figure}
\begin{figure}[!ht]
  \centering
   \includegraphics[width=\linewidth]{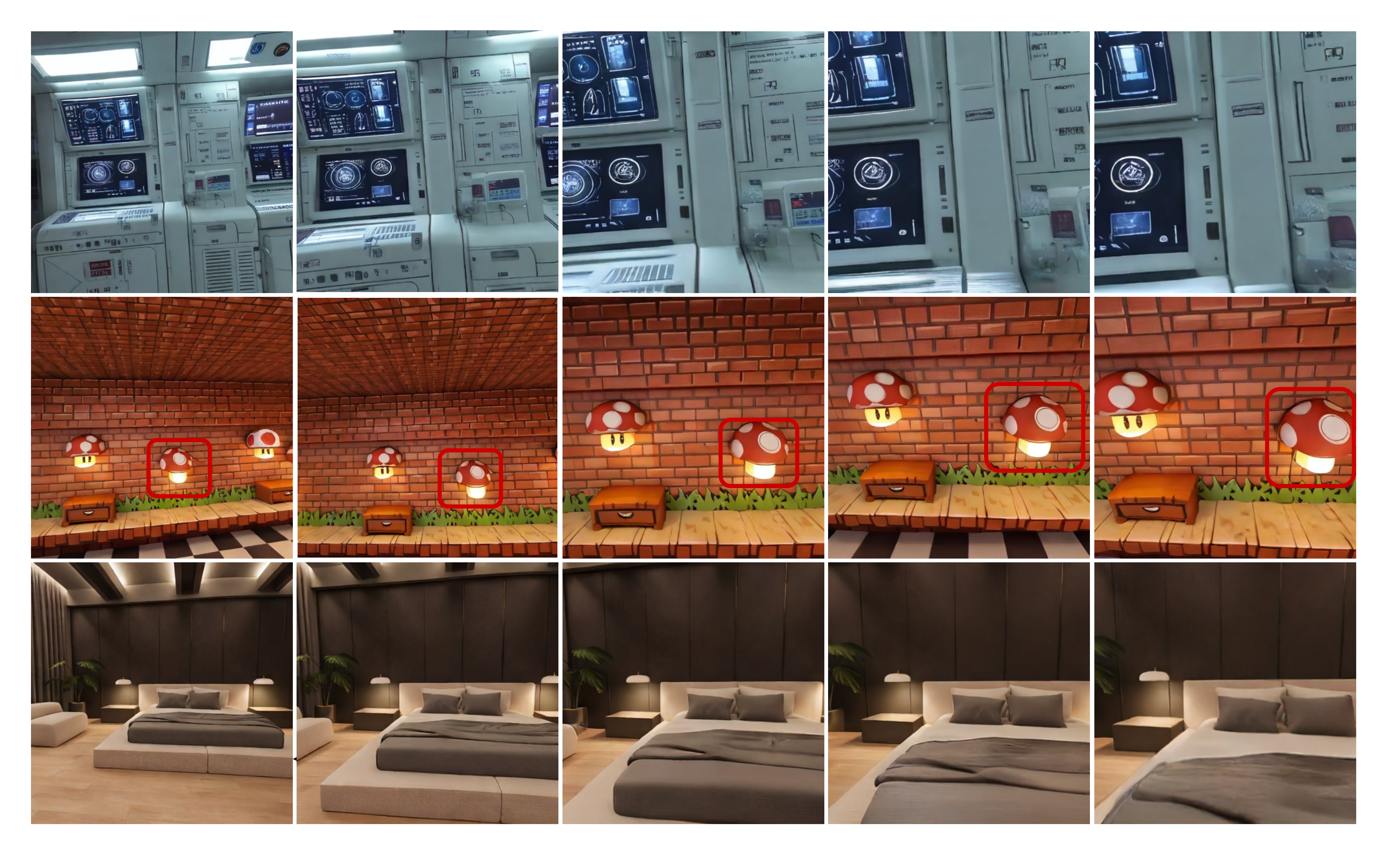}   
   \caption{Qualitative results of InSpatio-WorldFM. Across observations at varying viewing distances, the content and fine details generated by the FrameModel remain consistent.}   
   \label{fig:results4}   
   \vspace{-5mm}
\end{figure}

\subsection{Post-Training}

The goal of post-training is to transform the multi-step frame model from Stage II into an efficient generator capable of real-time interactive rendering.
We achieve this through Distribution Matching Distillation (DMD)~\citep{yin2024one}, which distills the multi-step diffusion model into a few-step generator with minimal loss in spatial consistency and visual fidelity.

\textbf{Distribution Matching Distillation.}
DMD trains a few-step diffusion student to match the output distribution of a pretrained multi-step diffusion model.
The core idea is to minimize an approximate KL divergence between the real distribution (defined by the base model) and the synthetic distribution (produced by the generator).
This is achieved by maintaining two diffusion models --- a frozen copy of the base model estimating the real score, and a dynamically-updated model continually trained on generator outputs estimating the fake score --- and using the difference between their denoising predictions as the gradient signal to update the generator.
A complementary regression loss on pre-computed noise-image pairs from the base model's deterministic sampler stabilizes training and preserves mode coverage. Extending the idea of Variational Score Distillation (VSD), DMD enables few-step inference acceleration while maintaining generation quality.

\textbf{Adaptation for InSpatio-WorldFM.}
We apply DMD to distill our middle-trained frame model into a few-step generator.
Through extensive experimentation, we arrive at two key findings:

\begin{itemize}
\item \textit{Two-step denoising outperforms one-step.} While DMD was originally designed for one-step generation, we find that a 2-step denoising schedule produces noticeably sharper details and better preserves fine-grained textures. Single-step denoising from pure noise can recover coarse geometric structure effectively but struggles to reconstruct fine-grained details in a single forward pass. The additional denoising step provides a dedicated refinement phase that addresses this limitation.

\item \textit{Intermediate timestep selection is critical.} In the 2-step schedule, the first step denoises from $t = T$ to an intermediate timestep $t_{\text{mid}}$, and the second step denoises from $t_{\text{mid}}$ to $t = 0$. If $t_{\text{mid}}$ is set too large (i.e., the first step terminates at a still high-noise state), the second step must perform single-step denoising from a high noise level --- encountering the same difficulty as 1-step generation in recovering fine details. Through systematic evaluation, we find that $t_{\text{mid}} = 200$ (on a 1000-step schedule) yields the best balance: the first step handles the majority of denoising and establishes the coarse spatial structure, while the second step refines from a relatively clean state where fine detail recovery is effective.
\end{itemize}

The distilled model maintains spatial consistency and visual fidelity over extended interactive sessions, producing temporally coherent frames without noticeable drift even during prolonged exploration.

\section{Evaluation}
We assess the spatial consistency and generation quality of InSpatio-WorldFM by examining the visual outputs from both the middle-trained fundamental frame model and the post-trained distilled variant across diverse scene types. 

As illustrated in Fig.~\ref{fig:results0}, each example consists of a reference image alongside a sequence of 10 frames rendered from different camera viewpoints. The fundamental frame model exhibits strong multi-view coherence, maintaining geometric structure and appearance consistency across substantial viewpoint changes. The generated frames preserve fine-grained details while adapting naturally to novel perspectives, demonstrating the model's capacity to reason about 3D spatial relationships.

To enable real-time interaction, we apply distribution matching distillation to obtain a distilled model for low-latency deployment.
In its baseline configuration at 512×512 resolution, InSpatio-WorldFM achieves around 25 FPS on a single H-series GPU through engineering optimizations such as KV-cache management and efficient VAE latent caching. Moreover, due to its low GPU memory footprint, the model can run at 10 FPS on an RTX 4090.
While distillation introduces a quality-speed trade-off, the perceptual difference remains minimal in practice. Visual comparison between the fundamental model and the distilled variant reveals that InSpatio-WorldFM successfully retains spatial coherence and geometric accuracy. The accelerated model continues to generate plausible content across diverse camera trajectories without noticeable degradation in structural consistency or the emergence of visual artifacts, confirming that our distillation approach effectively balances real-time performance with generation fidelity.

In long rendering tests under different historical frame contexts, InSpatio-WorldFM maintains consistent geometry and fine details across varying viewing distances, demonstrating the effectiveness of the memory strategy in guiding generation.

\section{Discussion and Conclusions}
In this report, we introduce InSpatio-WorldFM, an open-source real-time generative frame model designed for spatial intelligence. By adopting a frame-based generation paradigm and incorporating explicit 3D anchors with implicit spatial memory, the model maintains multi-view spatial consistency while enabling low-latency real-time inference.
\subsection{Limitations}
Despite the progress made, achieving an immersive real-time generative world model still faces several challenges.
\begin{itemize}

\item \textbf{Stable generation of dynamic content.} Both the frame-based model and the multi-view consistency training data contain limited dynamic content, which makes it difficult for the model to generate dynamic scenes with high quality and stability.

\item \textbf{Limited motion boundary.} The current historical memory relies on multi-view consistent observations or panoramic observations. However, these generation models suffer from high computational complexity and substantial memory consumption, restricting them to offline operation, which inevitably introduces motion boundaries during online inference.

\item \textbf{Interactive visual stability} The frame-based modeling strategy reduces interaction latency and improves responsiveness. However, due to the lack of temporal constraints between consecutive frames, the visual stability is limited, leading to noticeable frame jitter during interaction.

\end{itemize}

\subsection{Future Work}
Real-time spatial inference still has significant room for improvement. The frame-based inference architecture can benefit from many mature acceleration techniques, such as linear attention, efficient caching mechanisms, and various VAE optimizations, to further reduce computational cost and improve FPS. These improvements would enable spatial inference to be performed more efficiently on edge devices.  Using Gaussian Splatting (GS) primitives as 3D anchors could further enhance visual fidelity and reflective effects. Beyond efficiency and visual fidelity, we will focus on two key directions:
(1) \textbf{improving the generation of dynamic content}, and 
(2) \textbf{real-time expansion of the generation range}. 
These efforts will provide the foundation for efficient real-world modeling and spatial inference at effectively unbounded scales.

\newpage









{
    \small    
    \bibliographystyle{plain}
    \bibliography{main}

@String(CVPR= {IEEE Conf. Comput. Vis. Pattern Recog.})

@String(ICCV= {Int. Conf. Comput. Vis.})

@String(TOG= {ACM Trans. Graph.})

@String(AAAI = {AAAI})

@String(CVPR  = {CVPR})

@String(ICCV  = {ICCV})

@String(TOG   = {ACM TOG})

@misc{ue,
  author = {{Epic Games}},
  title = {{Unreal Engine}},
  year = {2023},
  howpublished = {\url{https://www.unrealengine.com/}},
  note = {Accessed: 2026-01-25}
}

@article{brooks2024video,
  title={Video generation models as world simulators},
  author={Tim Brooks and Bill Peebles and Connor Holmes and Will DePue and Yufei Guo and Li Jing and David Schnurr and Joe Taylor and Troy Luhman and Eric Luhman and Clarence Ng and Ricky Wang and Aditya Ramesh},
  journal={OpenAI Blog},
  year={2024}
}

@article{wan2025,
  title={Wan: Open and Advanced Large-Scale Video Generative Models},
  author={Wan Team},
  journal = {arXiv preprint arXiv:2503.20314},
  year={2025}
}

@article{kong2024hunyuanvideo,
  title={Hunyuanvideo: A systematic framework for large video generative models},
  author={Hunyuan Foundation Model Team},
  journal={arXiv preprint arXiv:2412.03603},
  year={2024}
}

@article{team2025hunyuanworld,
  title={Hunyuanworld 1.0: Generating immersive, explorable, and interactive 3d worlds from words or pixels},
  author={Team, HunyuanWorld and Wang, Zhenwei and Liu, Yuhao and Wu, Junta and Gu, Zixiao and Wang, Haoyuan and Zuo, Xuhui and Huang, Tianyu and Li, Wenhuan and Zhang, Sheng and others},
  journal={arXiv preprint arXiv:2507.21809},
  year={2025}
}

@article{yang2024cogvideox,
  title={Cogvideox: Text-to-video diffusion models with an expert transformer},
  author={Yang, Zhuoyi and Teng, Jiayan and Zheng, Wendi and Ding, Ming and Huang, Shiyu and Xu, Jiazheng and Yang, Yuanming and Hong, Wenyi and Zhang, Xiaohan and Feng, Guanyu and others},
  journal={arXiv preprint arXiv:2408.06072},
  year={2024}
}

@article{blattmann2023stable,
  title={Stable video diffusion: Scaling latent video diffusion models to large datasets},
  author={Andreas Blattmann and Tim Dockhorn and Sumith Kulal and Daniel Mendelevitch and Maciej Kilian and Dominik Lorenz and Yam Levi and Zion English and Vikram Voleti and Adam Letts and Varun Jampani and Robin Rombach},
  journal={arXiv preprint arXiv:2311.15127},
  year={2023}
}

@inproceedings{peebles2023scalable,
  title={Scalable diffusion models with transformers},
  author={Peebles, William and Xie, Saining},
  booktitle=ICCV,
  year={2023}
}

@inproceedings{bar2024lumiere,
  title={Lumiere: A space-time diffusion model for video generation},
  author={Omer Bar-Tal and Hila Chefer and Omer Tov and Charles Herrmann and Roni Paiss and Shiran Zada and Ariel Ephrat and Junhwa Hur and Guanghui Liu and Amit Raj and Yuanzhen Li and Michael Rubinstein and Tomer Michaeli and Oliver Wang and Deqing Sun and Tali Dekel and Inbar Mosseri},
  booktitle={SIGGRAPH Asia},
  year={2024}
}

@article{singer2022make,
  title={Make-a-video: Text-to-video generation without text-video data},
  author={Uriel Singer and Adam Polyak and Thomas Hayes and Xi Yin and Jie An and Songyang Zhang and Qiyuan Hu and Harry Yang and Oron Ashual and Oran Gafni and Devi Parikh and Sonal Gupta and Yaniv Taigman},
  journal={arXiv preprint arXiv:2209.14792},
  year={2022}
}

@article{lingbot-world,
      title={Advancing Open-source World Models}, 
      author={Robbyant Team and Zelin Gao and Qiuyu Wang and Yanhong Zeng and Jiapeng Zhu and Ka Leong Cheng and Yixuan Li and Hanlin Wang and Yinghao Xu and Shuailei Ma and Yihang Chen and Jie Liu and Yansong Cheng and Yao Yao and Jiayi Zhu and Yihao Meng and Kecheng Zheng and Qingyan Bai and Jingye Chen and Zehong Shen and Yue Yu and Xing Zhu and Yujun Shen and Hao Ouyang},
      journal={arXiv preprint arXiv:2601.20540},
      year={2026}
}

@article{huang2025voyager,
  title={Voyager: Long-range and world-consistent video diffusion for explorable 3d scene generation},
  author={Huang, Tianyu and Zheng, Wangguandong and Wang, Tengfei and Liu, Yuhao and Wang, Zhenwei and Wu, Junta and Jiang, Jie and Li, Hui and Lau, Rynson and Zuo, Wangmeng and others},
  journal={ACM Transactions on Graphics (TOG)},
  volume={44},
  number={6},
  pages={1--15},
  year={2025},
  publisher={ACM New York, NY, USA}
}

@article{wu2025video,
  title={Video world models with long-term spatial memory},
  author={Wu, Tong and Yang, Shuai and Po, Ryan and Xu, Yinghao and Liu, Ziwei and Lin, Dahua and Wetzstein, Gordon},
  journal={arXiv preprint arXiv:2506.05284},
  year={2025}
}

@inproceedings{ren2025gen3c,
  title={Gen3c: 3d-informed world-consistent video generation with precise camera control},
  author={Ren, Xuanchi and Shen, Tianchang and Huang, Jiahui and Ling, Huan and Lu, Yifan and Nimier-David, Merlin and M{\"u}ller, Thomas and Keller, Alexander and Fidler, Sanja and Gao, Jun},
  booktitle={Proceedings of the IEEE/CVF Conference on Computer Vision and Pattern Recognition},
  pages={6121--6132},
  year={2025}
}

@article{zhang2025matrix,
  title={Matrix-Game: Interactive World Foundation Model},
  author={Yifan Zhang and Chunli Peng and Boyang Wang and Puyi Wang and Qingcheng Zhu and Fei Kang and Biao Jiang and Zedong Gao and Eric Li and Yang Liu and Yahui Zhou},
  journal={arXiv preprint arXiv:2506.18701},
  year={2025}
}

@article{ding2025understanding,
  title={Understanding world or predicting future? a comprehensive survey of world models},
  author={Jingtao Ding and Yunke Zhang and Yu Shang and Jie Feng and Yuheng Zhang and Zefang Zong and Yuan Yuan and Hongyuan Su and Nian Li and Jinghua Piao and Yucheng Deng and Nicholas Sukiennik and Chen Gao and Fengli Xu and Yong Li},
  journal={ACM Computing Surveys},
  year={2025},
}

@inproceedings{yin2024one,
  title={One-step diffusion with distribution matching distillation},
  author={Yin, Tianwei and Gharbi, Micha{\"e}l and Zhang, Richard and Shechtman, Eli and Durand, Fredo and Freeman, William T and Park, Taesung},
  booktitle=CVPR,
  year={2024}
}

@inproceedings{chen2024pixart,
  title={Pixart-$\sigma$: Weak-to-strong training of diffusion transformer for 4k text-to-image generation},
  author={Chen, Junsong and Ge, Chongjian and Xie, Enze and Wu, Yue and Yao, Lewei and Ren, Xiaozhe and Wang, Zhongdao and Luo, Ping and Lu, Huchuan and Li, Zhenguo},
  booktitle={European Conference on Computer Vision},
  pages={74--91},
  year={2024},
  organization={Springer}
}

@article{li2025cameras,
  title={Cameras as relative positional encoding},
  author={Li, Ruilong and Yi, Brent and Liu, Junchen and Gao, Hang and Ma, Yi and Kanazawa, Angjoo},
  journal={arXiv preprint arXiv:2507.10496},
  year={2025}
}

@inproceedings{bai2025recammaster,
  title={{ReC}ammaster: Camera-controlled generative rendering from a single video},
  author={Bai, Jianhong and Xia, Menghan and Fu, Xiao and Wang, Xintao and Mu, Lianrui and Cao, Jinwen and Liu, Zuozhu and Hu, Haoji and Bai, Xiang and Wan, Pengfei and others},
  booktitle={Proceedings of the IEEE/CVF International Conference on Computer Vision},
  pages={14834--14844},
  year={2025}
}

@article{keetha2025mapanything,
  title={{MapA}nything: Universal feed-forward metric 3d reconstruction},
  author={Keetha, Nikhil and M{\"u}ller, Norman and Sch{\"o}nberger, Johannes and Porzi, Lorenzo and Zhang, Yuchen and Fischer, Tobias and Knapitsch, Arno and Zauss, Duncan and Weber, Ethan and Antunes, Nelson and others},
  journal={arXiv preprint arXiv:2509.13414},
  year={2025}
}

@article{zhou2018stereo,
  title={Stereo magnification: Learning view synthesis using multiplane images},
  author={Zhou, Tinghui and Tucker, Richard and Flynn, John and Fyffe, Graham and Snavely, Noah},
  journal={arXiv preprint arXiv:1805.09817},
  year={2018}
}

@inproceedings{rombach2022high,
  title={High-resolution image synthesis with latent diffusion models},
  author={Rombach, Robin and Blattmann, Andreas and Lorenz, Dominik and Esser, Patrick and Ommer, Bj{\"o}rn},
  booktitle={Proceedings of the IEEE/CVF conference on computer vision and pattern recognition},
  pages={10684--10695},
  year={2022}
}

@misc{labs2025flux1kontextflowmatching,
      title={{FLUX}.1 Kontext: Flow Matching for In-Context Image Generation and Editing in Latent Space},
      author={Black Forest Labs and Stephen Batifol and Andreas Blattmann and Frederic Boesel and Saksham Consul and Cyril Diagne and Tim Dockhorn and Jack English and Zion English and Patrick Esser and Sumith Kulal and Kyle Lacey and Yam Levi and Cheng Li and Dominik Lorenz and Jonas Müller and Dustin Podell and Robin Rombach and Harry Saini and Axel Sauer and Luke Smith},
      year={2025},
      eprint={2506.15742},
      archivePrefix={arXiv},
      primaryClass={cs.GR},
      url={https://arxiv.org/abs/2506.15742},
}

@misc{flux2024,
    author={Black Forest Labs},
    title={FLUX},
    year={2024},
    howpublished={\url{https://github.com/black-forest-labs/flux}},
}

@inproceedings{ling2024dl3dv,
  title={{DL3DV-10K}: A large-scale scene dataset for deep learning-based 3d vision},
  author={Ling, Lu and Sheng, Yichen and Tu, Zhi and Zhao, Wentian and Xin, Cheng and Wan, Kun and Yu, Lantao and Guo, Qianyu and Yu, Zixun and Lu, Yawen and others},
  booktitle={Proceedings of the IEEE/CVF Conference on Computer Vision and Pattern Recognition},
  pages={22160--22169},
  year={2024}
}

@inproceedings{yao2020blendedmvs,
  title={{BlendedMVS}: A large-scale dataset for generalized multi-view stereo networks},
  author={Yao, Yao and Luo, Zixin and Li, Shiwei and Zhang, Jingyang and Ren, Yufan and Zhou, Lei and Fang, Tian and Quan, Long},
  booktitle={Proceedings of the IEEE/CVF conference on computer vision and pattern recognition},
  pages={1790--1799},
  year={2020}
}

@misc{rtfm2025,
    author={WorldLabs},
    title={{RTFM}: A Real-Time Frame Model},
    year={2025},
    howpublished={\url{https://www.worldlabs.ai/blog/rtfm}},
}

@misc{genie3,
    author={Google DeepMind},
    title={Genie 3: A new frontier for world models},
    year={2025},
    howpublished={\url{https://deepmind.google/blog/genie-3-a-new-frontier-for-world-models/}},
}

@inproceedings{he2025cameractrl,
  title={Cameractrl ii: Dynamic scene exploration via camera-controlled video diffusion models},
  author={He, Hao and Yang, Ceyuan and Lin, Shanchuan and Xu, Yinghao and Wei, Meng and Gui, Liangke and Zhao, Qi and Wetzstein, Gordon and Jiang, Lu and Li, Hongsheng},
  booktitle={Proceedings of the IEEE/CVF International Conference on Computer Vision},
  pages={13416--13426},
  year={2025}
}

@article{agarwal2025cosmos,
  title={Cosmos world foundation model platform for physical ai},
  author={Agarwal, Niket and Ali, Arslan and Bala, Maciej and Balaji, Yogesh and Barker, Erik and Cai, Tiffany and Chattopadhyay, Prithvijit and Chen, Yongxin and Cui, Yin and Ding, Yifan and others},
  journal={arXiv preprint arXiv:2501.03575},
  year={2025}
}

@article{alhaija2025cosmos,
  title={Cosmos-transfer1: Conditional world generation with adaptive multimodal control},
  author={Alhaija, Hassan Abu and Alvarez, Jose and Bala, Maciej and Cai, Tiffany and Cao, Tianshi and Cha, Liz and Chen, Joshua and Chen, Mike and Ferroni, Francesco and Fidler, Sanja and others},
  journal={arXiv preprint arXiv:2503.14492},
  year={2025}
}

@inproceedings{wang2024dust3r,
  title={Dust3{R}: Geometric 3d vision made easy},
  author={Wang, Shuzhe and Leroy, Vincent and Cabon, Yohann and Chidlovskii, Boris and Revaud, Jerome},
  booktitle={Proceedings of the IEEE/CVF conference on computer vision and pattern recognition},
  pages={20697--20709},
  year={2024}
}

@inproceedings{wang2025vggt,
  title={{VGGT}: Visual geometry grounded transformer},
  author={Wang, Jianyuan and Chen, Minghao and Karaev, Nikita and Vedaldi, Andrea and Rupprecht, Christian and Novotny, David},
  booktitle={Proceedings of the Computer Vision and Pattern Recognition Conference},
  pages={5294--5306},
  year={2025}
}

@inproceedings{zhou2025scenex,
  title={Scene{X}: Procedural controllable large-scale scene generation},
  author={Zhou, Mengqi and Wang, Yuxi and Hou, Jun and Zhang, Shougao and Li, Yiwei and Luo, Chuanchen and Peng, Junran and Zhang, Zhaoxiang},
  booktitle={Proceedings of the AAAI Conference on Artificial Intelligence},
  volume={39},
  number={10},
  pages={10806--10814},
  year={2025}
}

@inproceedings{raistrick2024infinigen,
  title={Infinigen indoors: Photorealistic indoor scenes using procedural generation},
  author={Raistrick, Alexander and Mei, Lingjie and Kayan, Karhan and Yan, David and Zuo, Yiming and Han, Beining and Wen, Hongyu and Parakh, Meenal and Alexandropoulos, Stamatis and Lipson, Lahav and others},
  booktitle={Proceedings of the IEEE/CVF Conference on Computer Vision and Pattern Recognition},
  pages={21783--21794},
  year={2024}
}

@inproceedings{yang2025layerpano3d,
  title={Layerpano{3D}: Layered 3d panorama for hyper-immersive scene generation},
  author={Yang, Shuai and Tan, Jing and Zhang, Mengchen and Wu, Tong and Wetzstein, Gordon and Liu, Ziwei and Lin, Dahua},
  booktitle={Proceedings of the special interest group on computer graphics and interactive techniques conference conference papers},
  pages={1--10},
  year={2025}
}

@article{chung2023luciddreamer,
  title={{LucidD}reamer: Domain-free generation of {3D} gaussian splatting scenes},
  author={Chung, Jaeyoung and Lee, Suyoung and Nam, Hyeongjin and Lee, Jaerin and Lee, Kyoung Mu},
  journal={arXiv preprint arXiv:2311.13384},
  year={2023}
}

@inproceedings{yu2024wonderjourney,
  title={{WonderJ}ourney: Going from anywhere to everywhere},
  author={Yu, Hong-Xing and Duan, Haoyi and Hur, Junhwa and Sargent, Kyle and Rubinstein, Michael and Freeman, William T and Cole, Forrester and Sun, Deqing and Snavely, Noah and Wu, Jiajun and others},
  booktitle={Proceedings of the IEEE/CVF Conference on Computer Vision and Pattern Recognition},
  pages={6658--6667},
  year={2024}
}

@inproceedings{yu2025wonderworld,
  title={{WonderW}orld: Interactive 3d scene generation from a single image},
  author={Yu, Hong-Xing and Duan, Haoyi and Herrmann, Charles and Freeman, William T and Wu, Jiajun},
  booktitle={Proceedings of the Computer Vision and Pattern Recognition Conference},
  pages={5916--5926},
  year={2025}
}

@article{liu2024physics3d,
  title={Physics3{D}: Learning physical properties of {3D} gaussians via video diffusion},
  author={Liu, Fangfu and Wang, Hanyang and Yao, Shunyu and Zhang, Shengjun and Zhou, Jie and Duan, Yueqi},
  journal={arXiv preprint arXiv:2406.04338},
  year={2024}
}

@inproceedings{zhang2024physdreamer,
  title={Phys{D}reamer: Physics-based interaction with 3d objects via video generation},
  author={Zhang, Tianyuan and Yu, Hong-Xing and Wu, Rundi and Feng, Brandon Y and Zheng, Changxi and Snavely, Noah and Wu, Jiajun and Freeman, William T},
  booktitle={European Conference on Computer Vision},
  pages={388--406},
  year={2024},
  organization={Springer}
}

@article{gao2024cat3d,
  title={Cat3{D}: Create anything in 3d with multi-view diffusion models},
  author={Gao, Ruiqi and Holynski, Aleksander and Henzler, Philipp and Brussee, Arthur and Martin-Brualla, Ricardo and Srinivasan, Pratul and Barron, Jonathan T and Poole, Ben},
  journal={arXiv preprint arXiv:2405.10314},
  year={2024}
}

@inproceedings{zhai2025stargen,
  title={Star{G}en: A spatiotemporal autoregression framework with video diffusion model for scalable and controllable scene generation},
  author={Zhai, Shangjin and Ye, Zhichao and Liu, Jialin and Xie, Weijian and Hu, Jiaqi and Peng, Zhen and Xue, Hua and Chen, Danpeng and Wang, Xiaomeng and Yang, Lei and others},
  booktitle={Proceedings of the Computer Vision and Pattern Recognition Conference},
  pages={26822--26833},
  year={2025}
}

@misc{wang2024moge,
    title={{MoGe}: Unlocking Accurate Monocular Geometry Estimation for Open-Domain Images with Optimal Training Supervision},
    author={Wang, Ruicheng and Xu, Sicheng and Dai, Cassie and Xiang, Jianfeng and Deng, Yu and Tong, Xin and Yang, Jiaolong},
    year={2024},
    eprint={2410.19115},
    archivePrefix={arXiv},
    primaryClass={cs.CV},
    url={https://arxiv.org/abs/2410.19115}, 
}

@inproceedings{li2018megadepth,
  title={{MegaDepth}: Learning single-view depth prediction from internet photos},
  author={Li, Zhengqi and Snavely, Noah},
  booktitle={Proceedings of the IEEE conference on computer vision and pattern recognition},
  pages={2041--2050},
  year={2018}
}

@article{feng2023diffusion360,
  title={Diffusion360: Seamless 360 degree panoramic image generation based on diffusion models},
  author={Feng, Mengyang and Liu, Jinlin and Cui, Miaomiao and Xie, Xuansong},
  journal={arXiv preprint arXiv:2311.13141},
  year={2023}
}

@inproceedings{zhang2024taming,
  title={Taming stable diffusion for text to 360 panorama image generation},
  author={Zhang, Cheng and Wu, Qianyi and Gambardella, Camilo Cruz and Huang, Xiaoshui and Phung, Dinh and Ouyang, Wanli and Cai, Jianfei},
  booktitle={Proceedings of the IEEE/CVF Conference on Computer Vision and Pattern Recognition},
  pages={6347--6357},
  year={2024}
}

@inproceedings{zhou2025stable,
  title={Stable virtual camera: Generative view synthesis with diffusion models},
  author={Zhou, Jensen and Gao, Hang and Voleti, Vikram and Vasishta, Aaryaman and Yao, Chun-Han and Boss, Mark and Torr, Philip and Rupprecht, Christian and Jampani, Varun},
  booktitle={Proceedings of the IEEE/CVF International Conference on Computer Vision},
  pages={12405--12414},
  year={2025}
}

@article{sun2025worldplay,
  title={Worldplay: Towards long-term geometric consistency for real-time interactive world modeling},
  author={Sun, Wenqiang and Zhang, Haiyu and Wang, Haoyuan and Wu, Junta and Wang, Zehan and Wang, Zhenwei and Wang, Yunhong and Zhang, Jun and Wang, Tengfei and Guo, Chunchao},
  journal={arXiv preprint arXiv:2512.14614},
  year={2025}
}

@article{videoworldsimulators2024,
  title={Video generation models as world simulators},
  author={Tim Brooks and Bill Peebles and Connor Holmes and Will DePue and Yufei Guo and Li Jing and David Schnurr and Joe Taylor and Troy Luhman and Eric Luhman and Clarence Ng and Ricky Wang and Aditya Ramesh},
  year={2024},
  url={https://openai.com/research/video-generation-models-as-world-simulators},
}
}

\end{document}